%% file: main.tex
\documentclass[10pt,twocolumn,letterpaper]{article}

\usepackage{cvpr}      

\usepackage{color}

\definecolor{cvprblue}{rgb}{0.21,0.49,0.74}
\usepackage[pagebackref,breaklinks,colorlinks,allcolors=cvprblue]{hyperref}
\usepackage{graphicx}
\usepackage[utf8]{inputenc}
\usepackage{caption}
\usepackage{amsmath}
\usepackage{amsthm}
\usepackage{booktabs}
\usepackage{algorithm}
\usepackage{algorithmic}
\usepackage[switch]{lineno}
\usepackage{multirow}
\usepackage{bbding}
\usepackage{xcolor} 
\usepackage{xspace}
\usepackage{makecell}
\usepackage{colortbl}
\usepackage{breakurl}

\newcommand{\dataset}{FunBench\xspace}

\newcommand{\ve}[1]{\textcolor{blue}{\texttt{#1}}}
\newcommand{\llm}[1]{\textcolor{purple}{\texttt{#1}}}

\def\paperID{*****} 
\def\confName{CVPR}
\def\confYear{2025}

\title{FunBench: Benchmarking Fundus Reading Skills of MLLMs}

\author{Qijie Wei\quad
Kaiheng Qian\quad
Xirong Li\thanks{Corresponding author, xirong@ruc.edu.cn}\\
{Renmin University of China} \\
{https://huggingface.co/datasets/AIMClab-RUC/FunBench}
}

\begin{document}
\maketitle

\begin{abstract}
Multimodal Large Language Models (MLLMs) have shown significant potential in medical image analysis. However, their capabilities in interpreting fundus images, a critical skill for ophthalmology, remain under-evaluated. Existing benchmarks lack fine-grained task divisions and fail to provide modular analysis of its two key modules, i.e., large language model (LLM) and vision encoder (VE). 
This paper introduces FunBench, a novel visual question answering (VQA) benchmark designed to comprehensively evaluate MLLMs' fundus reading skills. FunBench features a hierarchical task organization across four levels (modality perception, anatomy perception, lesion analysis, and disease diagnosis).
It also offers three targeted evaluation modes: linear-probe based VE evaluation, knowledge-prompted LLM evaluation, and holistic evaluation. Experiments on nine open-source MLLMs plus GPT-4o reveal significant deficiencies in fundus reading skills, particularly in basic tasks such as laterality recognition. The results highlight the limitations of current MLLMs and emphasize the need for domain-specific training and improved LLMs and VEs.     
\end{abstract}

\section{Introduction}
\input{introduction}
\section{\dataset Construction}

\input{benchmark_construction}

\section{Evaluating MLLMs on FunBench}
\input{experiments}

\section{Conclusions}

Our evaluation of varied MLLMs on the new FunBench benchmark supports conclusions as follows. 
First, the MLLMs evaluated remain weak for performing fundus reading tasks related to anatomy perception, lesion analysis and disease diagnosis. 
Second, they rely much more on their LLMs rather than their VEs. Third, the VEs are less effective than DINOv2.  
Lastly, the overall best performance of HuatuoGPT-Vision shows the importance of domain-specific training. The future design of the training procedure needs to consider the big picture, or we risk developing an MLLM that lacks basic fundus reading skills.

{
    \small
    \bibliographystyle{ieeenat_fullname}
    \bibliography{ref}
}

\end{document}

%% file: introduction.tex
\begin{figure*}[tb!]
    \centering
    
    \subfloat[Tasks at four levels, from \emph{modality perception} (L1) to \emph{disease diagnosis} (L4).]
    {\includegraphics[width=1\textwidth]{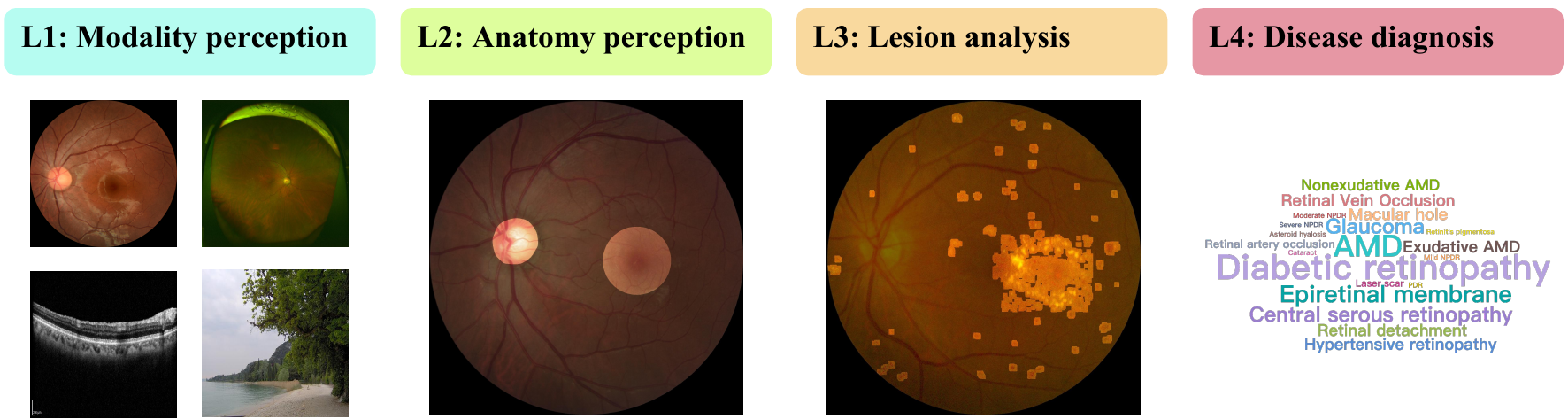}\label{fig:tasks}} 

    \subfloat[E-mode I]{
        \includegraphics[height=0.35\textwidth]{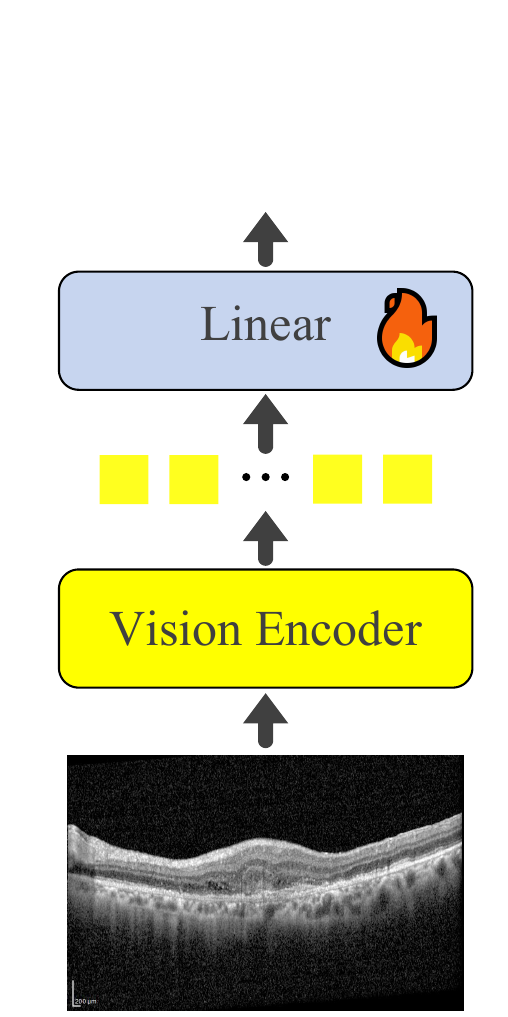}\label{fig:e-mode-1}} 
    \subfloat[E-mode II]
    {\includegraphics[height=0.35\textwidth]{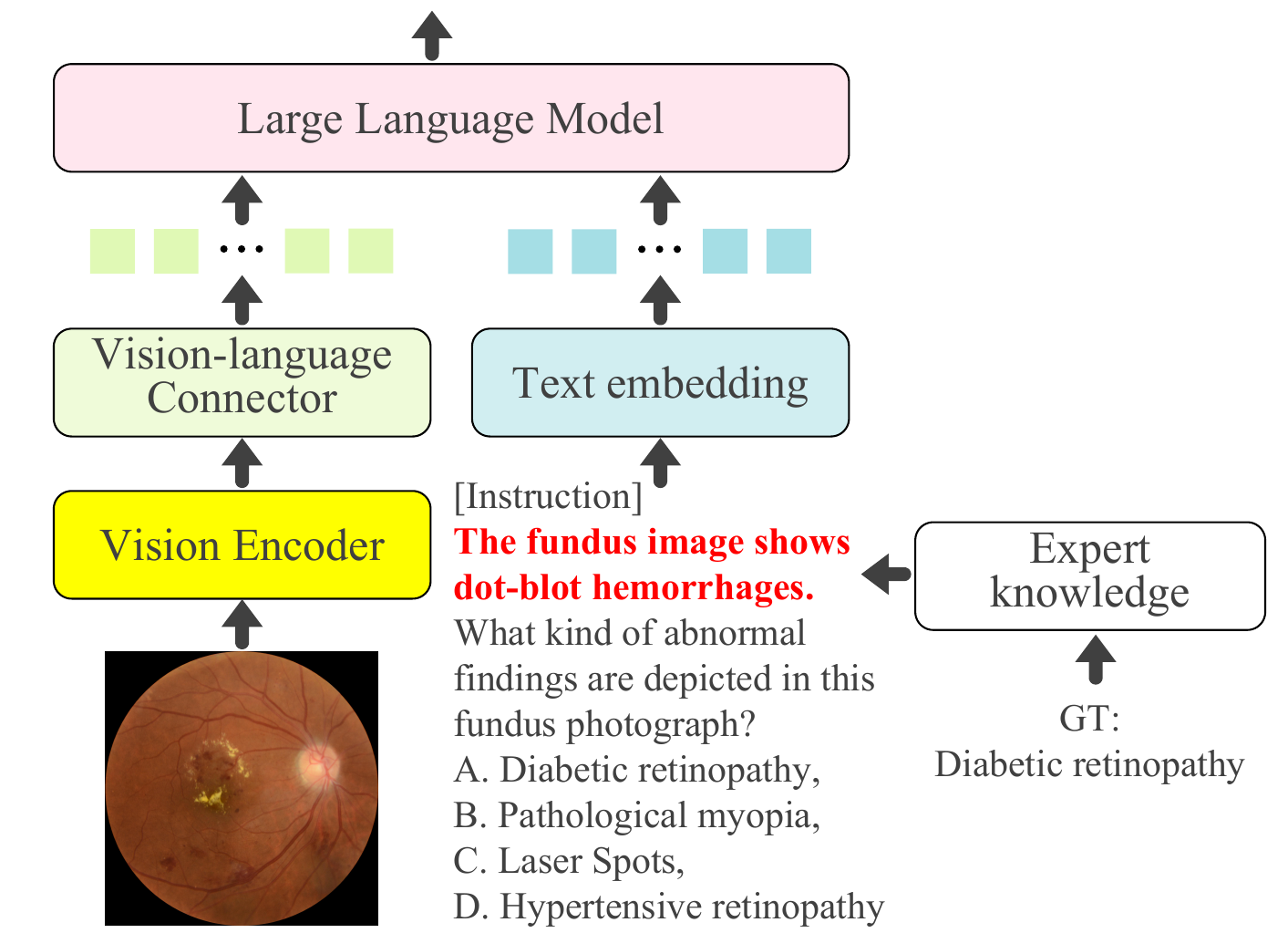}\label{fig:e-mode-2}} 
    \subfloat[E-mode III]
    {\includegraphics[height=0.35\textwidth]{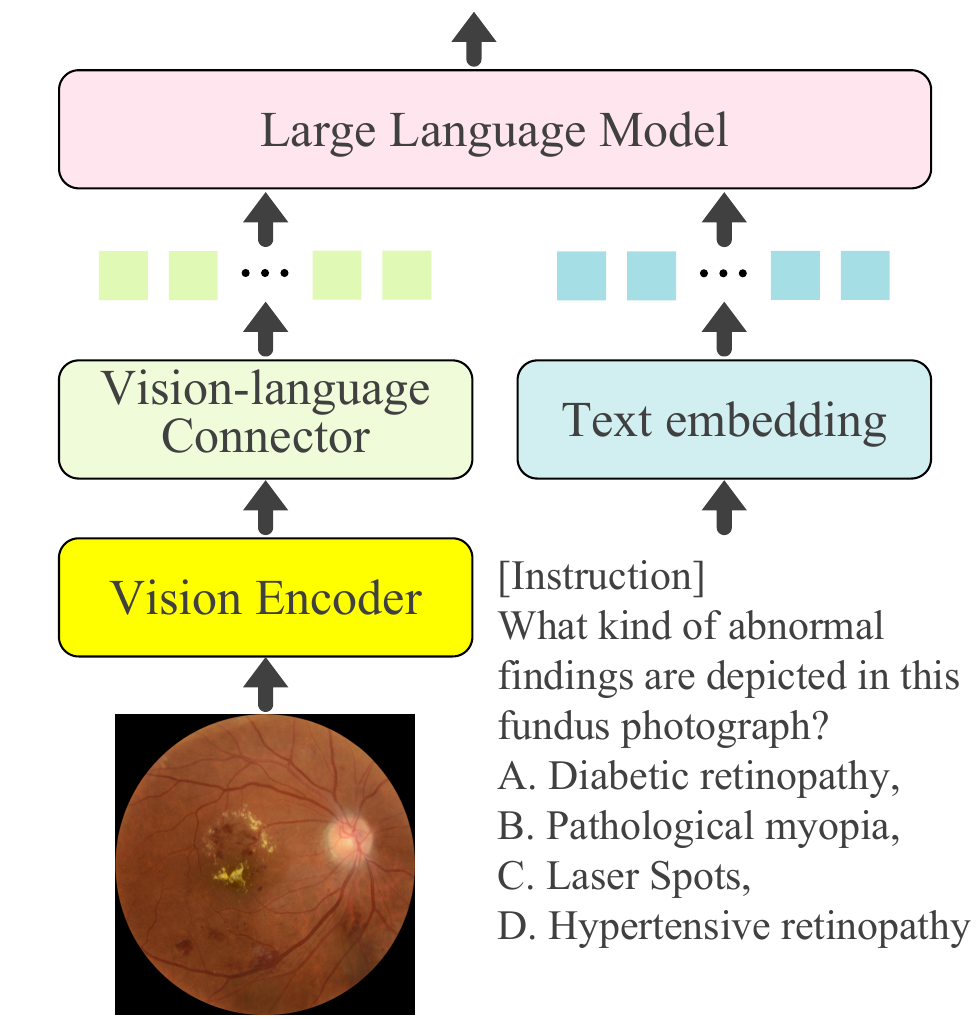}\label{fig:e-mode-3}} 
    \caption{\textbf{Proposed FunBench for assessing an MLLM's funding reading skills} by (a) varied-level tasks and three distinct evaluation modes, \ie (b) \emph{E-mode I}: linear-probe based vision encoder (\ve{VE}) evaluation, (c) \emph{E-mode II}: knowledge-prompted large language model (\llm{LLM}) evaluation and (d) \emph{E-mode III}: holistic evaluation.}
    \label{fig:framework}
\end{figure*}

Multimodal Large Language Models (MLLMs), with their strong capabilities in generic visual content understanding, are rocking the field of medical image analysis \cite{li2024llavamed,he2024pitvqa} and consequently reshaping the research landscape of medical image-based disease diagnosis \cite{deng2024ophglm,zhang2024incorporating}. Consider AI-enabled Ophthalmology for instance. The combination of remote MLLMs \emph{and} locally deployed non-invasive fundus imaging devices such as color fundus photography (CFP) makes high-quality primary diabetes care possible at community clinics \cite{DeepDR-LLM}. 
While research on medical MLLMs is growing rapidly \cite{DeepDR-LLM,deng2024ophglm,wang2024eyegraphgpt,yeh2024insight}, we observe that the development of ophthalmology-targeted benchmarks is lagging behind. This paper develops \textbf{FunBench}, a new benchmark for evaluating the efficacy of \emph{open-source} MLLMs for \textbf{fundus reading} tasks of varied difficulties, see Fig. \ref{fig:framework}. 

While medical benchmarks such as OmniMedVQA\cite{hu2024omnimedvqa} and GMAI-MMBench \cite{yegmai} have retinal images included, they treat retinal image-based visual question answering (VQA) as a \emph{single} task. A detailed and structured evaluation of how well a specific model can interpret retinal images is naturally absent from these \emph{general-purpose} benchmarks. Towards filling the gap, LMOD \cite{qin2024lmod} has been developed, evaluating the performance of MLLMs on recognizing major anatomical structures of the fundus, \eg optic cup, optic disc and fovea, and on recognizing two diseases, \ie glaucoma and macular hole. Therefore, LMOD enables a more detailed assessment as opposed to OmniMedVQA and GMAI-MMBench. Our FunBench technically differs from LMOD for its \emph{hierarchical} task organization and \emph{targeted} evaluation modes.  

The construction of FunBench is driven by our quest to answer two fundamental questions related to the assessment of a MLLM's fundus reading skills. That is, \emph{what to ask} and \emph{how to ask}. On answering the first question, we consider four levels of tasks, ranging from \emph{low-level} modality and anatomy perception to \emph{high-level} lesion analysis and disease diagnosis.
Such a task organization enables a comprehensive assessment of the level and extent to which an MLLM has mastered its fundus reading skills, an evaluation that prior benchmarks have not adequately supported. For answering the second question, our evaluation is not only targeted at the MLLM as a whole, but also considers its two key modules, \ie vision encoder (\ve{VE}) and large language model (\llm{LLM}). Such a design enables a joint evaluation that is both holistic and modular --  an analytical approach not used in the previous work \cite{hu2024omnimedvqa, qin2024lmod}.

To sum up, our contributions are three-fold as follows: \\
$\bullet$ \textbf{Dataset}. We introduce FunBench, a novel benchmark for evaluating fundus reading skills of MLLMs. \\
$\bullet$ \textbf{Evaluation}. We evaluate nine open-source MLLMs, released between 2023.10 to 2025.01 on HuggingFace. These models cover five \ve{VE}s and seven  \llm{LLM}s. We include GPT-4o as a proprietary baseline and DIVOv2 \cite{oquab2024dinov2} as a \ve{VE} baseline. \\
$\bullet$ \textbf{Findings}. The MLLMs evaluated rely heavily on the internal \llm{LLM}s to perform the fundus reading tasks. The models possess quite limited fundus reading skills. In particular, they lack some basic skills such as laterality recognition. 

%% file: benchmark_construction.tex


\subsection{Dataset Curation}\label{ssec:data}

\subsubsection{Hierarchical Task Organization}


Depending on the extent to which a professional fundus reading skill is required, we consider four levels of tasks, ranging from basic modality perception to advanced fundus image interpretation.
\\
$\bullet$ \textbf{Level 1 (L1): Modality perception}. A model possessing L1 skills shall identify the imaging technique used to produce a given fundus image. In a native setting, one might consider selecting ``fundus'' from multiple choices such as ``natural'', ``painting'', and ``remote sensing''. A more difficult setting is to select among varied fundus modalities  such as \texttt{CFP}, Optical coherence tomography (\texttt{OCT}) and ultra-wide field fundus photography (\texttt{UWF}) plus other medical imaging such as \texttt{X-ray} and magnetic resonance imaging (\texttt{MRI}). We name the two settings \emph{coarse-grained} modality perception (\textbf{L1a}) and \emph{fine-grained} modality perception (\textbf{L1b}), respectively. \\
$\bullet$ \textbf{Level 2 (L2): Anatomy perception}. The optic disc (OD) and the fovea are two major anatomical structures in the retina. Their  visual patterns are relatively clear: OD typically 
 appears as an oval bright object in a color fundus image, whilst the fovea is centered in the darkest area on OD's temporal side \cite{mlmi2019-uwf}. Moreover, the laterality of the fundus image, \ie whether from a left or right eye, can be determined by the relative position of the OD in the given image \cite{mmm2019-left-right-eye}. Hence, a model possessing L2 skills shall tell the relative OD-fovea position (\textbf{L2a}) and recognize the laterality  (\textbf{L2b}). \\
%
$\bullet$ \textbf{Level 3 (L3): Lesion analysis}. 
Lesions are pathological alterations caused by varied diseases and can be observed (to some extent) by specific fundus imaging techniques. Recognizing what the lesions are, where they occur, how large and how many they are essential for reliable and explainable disease diagnosis. Consider diabetic retinopathy (DR) grading for instance. A sufficient criterion for severe nonproliferative DR is the presence of over 20 haemorrhages in each of the nasal, temporal, superior, and inferior quadrants of the fundus \cite{retinal-lesions}. Therefore, a model mastering L3 skills shall be able to perform lesion \emph{recognition} (\textbf{L3a}), \emph{localization} (\textbf{L3b}), \emph{size estimation} (\textbf{L3c}) and \emph{counting} (\textbf{L3d}). \\
%
$\bullet$ \textbf{Level 4 (L4): Disease diagnosis}. 
Disease diagnosis typically requires making  judgments based on a comprehensive consideration of  the lesions presented, changes in anatomical structures and overall appearance about the fundus.
Such a requirement naturally places L4 skills at the highest level, which typically takes medical students years to master.




\subsubsection{Data Sources}

In order to instantiate the above four-level tasks, we adapt the following 14 public datasets: 1) six \texttt{CFP} datasets: IDRiD \cite{idrid}, DDR \cite{DDR}, JSIEC \cite{jsiec}, RFMiD \cite{rfmid}, OIA-ODIR \cite{oia-odir} and Retinal-Lesions \cite{retinal-lesions}, 2) five \texttt{OCT} datasets: OCTDL~\cite{octdl}, NEH~\cite{NEH}, OCTID~\cite{octid}, UCSD~\cite{UCSD}, and RETOUCH~\cite{retouch}, 3) 
one \texttt{UWF} dataset: TOP\footnote{\url{https://github.com/DateCazuki/Fundus\_Diagnosis}}, and 4) two multimodal: MMC-AMD (\texttt{CFP+OCT}) \cite{mmc-amd2} and DeepDRiD \cite{deepdrid} (\texttt{CFP+UWF}). 

Subject to their original annotations, the use of the datasets in specific tasks is listed in Table \ref{tab:datasets}. Note that we take from  each dataset its test split\footnote{In case no official data split is provided, we randomly select 20\% of the dataset.} to form FunBench, with the remaining part preserved as a development set for future usage,  \eg supervised fine-tuning.  
Provided with the four lesion-annotated datasets, \ie DDR, IDRiD, Retinal-Lesions, and RETOUCH, we subdivide each of the four L3 tasks by distinct lesions whenever applicable, resulting in 39 subtasks in total. Using the multiple disease-annotated datasets, we now instantiate the L4 skills with 4 concrete tasks, namely \emph{binary-condition (normal or abnormal) diagnosis} (\textbf{L4a}), \emph{multi-condition diagnosis} (\textbf{L4b}), \emph{DR grading} (\textbf{L4c}) and \emph{fine-grained age-related macular degeneration (AMD) categorization} (\textbf{L4d}).

\input{tables/dataset}

\subsubsection{From Annotations to VQA Quadruples}

Similar to OmniMedVQA \cite{hu2024omnimedvqa}, we generate multi-choice VQA quadruples of (\texttt{image}, \texttt{question}, \texttt{options}, \texttt{answer})  from given images and their associated labels by auto-completing a number of predefined \emph{task-specific} question templates. See samples in Table \ref{tab:datasets}.

To direct the MLLM to select directly from the provided options, we prepend specific instructions to each question. For single-answer questions, the instruction is ``\emph{Please choose the most suitable option based on the image and the question. Answer with the option's letter directly}''. For multi-answer questions, the instruction is ``\textit{Please choose all the suitable options based on the image and the question. Answer with the option's letter directly. Please separate the answers with commas if needed}.''.

\subsection{Targeted Evaluation Modes} \label{ssec:e-modes}

In order to assess a given MLLM and its two key modules, \ie \llm{LLM} and \ve{VE}, we present three targeted evaluation modes (E-mode) in a bottom-up manner. 

\textbf{E-Mode I: Linear-probe based VE Evaluation}.
To assess the effectiveness of the \ve{VE} in extracting visual features from a given fundus image, we employ the widely used linear probe (LP) technique  \cite{tu2023visual}. As illustrated in Fig. \ref{fig:e-mode-1}, LP trains a \texttt{Linear}-layer based classification header per (sub-)task using the task-specific development dataset.
As such, we omit tasks that cannot be directly tackled as a classification problem, \eg \textbf{L3b}, \textbf{L3c} and \textbf{L3d}, and tasks trivial for LP, \eg \textbf{L1a} and \textbf{L1b}. 
Comparing \ve{VE}s used by different MLLMs in this manner helps reveal which \ve{VE} is more suited for fundus feature extraction.


\textbf{E-Mode II: Knowledge-prompted LLM evaluation}. 
As the \llm{LLM} module has been re-trained to handle multimodal tokens, evaluating the module by directly submitting a textual question is problematic. To reduce the influence of the \ve{VE}, we propose a simple knowledge-prompted  evaluation strategy as follows. Given a test image and its associated task-specific label, we convert the label to an indirect description by querying an expert-knowledge database (EyeWiki). Such a description is further formatted in a task-specific manner. Consider \textbf{L2b} \emph{laterality recognition} for instance. A left-eye image will be described as ``\emph{fovea located to the right side of the optic disc}''. As for \emph{hard exudate recognition}, one of the subtasks of \textbf{L3a}, the corresponding description will be ``\emph{white or yellowish  deposits with sharp margins}''. As shown in Fig. \ref{fig:e-mode-2}, by placing the description before the question, we perform knowledge-prompted LLM evaluation. 
Note that for this evaluation mode, some of the tasks, \eg \textbf{L1}, \textbf{L2a} and \textbf{L4a}, will be omitted as the provided prompts would make the tasks trivial to accomplish.

\textbf{E-Mode III: Holistic Evaluation}. 
This mode offers an end-to-end evaluation of the MLLM, see Fig. \ref{fig:e-mode-3}. By submitting a multimodal multi-option question to the model followed by a string comparison between the model's answer and the ground truth, a binary output is obtained. We found in preliminary experiments that some MLLMs, \eg  Qilin-Med-VL-Chat \cite{qilin}, LLaVA-Med-v1.5-7B \cite{li2024llavamed} and Janus-Pro-7B  \cite{janus}, cannot follow our instruction that requires them to produce a single-character response. Instead, they tend to respond with more extensive, open-ended text. In order to select the option that best matches with such text, we perform text-to-text semantic matching by a pre-trained Sentence-BERT \cite{reimers-2019-sentence-bert}, which encodes a given sentence into a 384-d embedding vector.

\subsubsection{Performance Metrics}

An AI-assisted disease diagnosis system naturally aims for fewer missed detections and false alarms, which can be measured by \emph{Sensitivity} and \emph{Specificity}, respectively. We report their harmonic mean, \emph{a.k.a.} \emph{F1} score, as a combined metric. For a multi-class task such as DR grading (\textbf{L4c}), a task-level F1 is computed as the mean value of F1 scores across all its classes.


%% file: tables/dataset.tex
\begin{table*}[tb!]
    \caption{\textbf{Statistics of FunBench}: 16,348 fundus images and 91,810 visual questions \wrt 10 tasks in total.}
    \label{tab:datasets}
    \renewcommand{\arraystretch}{1}
    \centerline{
        \scalebox{0.74}{ 
            \begin{tabular}{@{}ll rr rr ll ll @{}}
               \toprule
                \multirow{2}{*}{\textbf{Level}} && \multicolumn{3}{c}{\textbf{\#Visual questions}} && \multirow{2}{*}{\textbf{Sample question}} && \multirow{2}{*}{\textbf{Data sources}} \\
                \cline{3-5}
                && \textit{Single-ans.}  && \textit{Multi-ans.}  \\
                \midrule
                \rowcolor[rgb]{0.796,0.996,0.965} \Gape[0pt][2pt]{\makecell[c]{L1\\ \#Tasks: 2}}    && 32,696 && 0 &&  \Gape[0pt][2pt]{\makecell[l]{What method was used to capture this image? \\A. Magnetic resonance imaging, \\B. Ultra-wide field fundus photography, \\C. Color fundus photography, \\D. Optical coherence tomography.}} && All datasets \\
                \rowcolor[rgb]{0.910,1.0,0.733} \Gape[0pt][2pt]{\makecell[c]{L2\\\#Tasks: 2}}    && 10,980 && 0 && \Gape[0pt][2pt]{\makecell[l]{Which eye is shown in this image, the left or the right?\\A. Right eye, \\ B. Left eye.}} && \Gape[0pt][2pt]{\makecell[l]{\texttt{[CFP]} DDR, DeepDRiD, IDRiD, \\ OIA-ODIR, Retinal-Lesions\\ \texttt{[UWF]} TOP\\ \texttt{[CFP+UWF]} DeepDRiD}} \\
                \rowcolor[rgb]{0.984,0.894,0.741} \Gape[0pt][2pt]{\makecell[c]{L3\\ \#Tasks: 4\\ \#Subtasks: 39}}  && 15,606  && 7,237 && \Gape[0pt][2pt]{\makecell[l]{What are the positions of the Haemorrhages in the fundus image?\\A. Nasal side of the optic disc center, \\B. Temporal side of the optic disc center,\\C. Superior side of the optic disc center, \\D. Inferior side of the optic disc center, \\E. Not observed in the image.}} && \Gape[0pt][2pt]{\makecell[l]{\texttt{[CFP]} DDR, IDRiD, Retinal-Lesion\\ \texttt{[OCT]} RETOUCH}}\\
                \rowcolor[rgb]{0.929,0.718,0.753} \Gape[0pt][2pt]{\makecell[c]{L4\\\#Tasks: 4}}    && 20,177 && 5,114 && \Gape[0pt][2pt]{\makecell[l]{Which abnormalities can be seen in this fundus image?\\A. Glaucoma, \\B. Diabetic retinopathy, \\C. No abnormality, \\D. Age-related macular degeneration, \\E. Hypertensive retinopathy.}}  && \Gape[0pt][2pt]{\makecell[l]{\texttt{[CFP]} DDR, IDRiD, OIA-ODIR, \\JSIEC, RFMiD, Retinal-Lesions\\ \texttt{[OCT]} NEH, OCTDL, OCTID, UCSD \\ \texttt{[UWF]} TOP \\ \texttt{[CFP+OCT]} MMC-AMD\\ \texttt{[CFP+UWF]} DeepDRiD}}\\

              \bottomrule
            \end{tabular}
        }
    }
\end{table*}

%% file: experiments.tex
\subsection{Choices of MLLMs}

For reproducible research, we focus on open-source MLLMs. Subject to our GPU computation capability, we select MLLMs at \textbf{7B}/\textbf{8B} scales, compiling a list of six general-purpose and three medical models, see Table \ref{tab:mllm_llm_ve}.
In addition, we include GPT-4o \cite{gpt4o} as a proprietary baseline\footnote{API version: gpt-4o-2024-08-06.}. 
We adopt DINOv2-large  \cite{oquab2024dinov2}, a strong vision foundation model, as a \ve{VE} baseline.


\input{tables/mllm_llm_ve}


\input{tables/result_overall}

\subsection{Results}


\textbf{VE Comparison}. The performance of the different \ve{VE}s is shown in the \texttt{E-mode I} part of Table \ref{tab:res_overall}. DINOv2 is the best, though not used by the MLLMs. By contrast, the most popular CLIP-ViT, with mean score of 0.578, has turned out to be the least effective. Checking its performance per task, we see the largest performance gap at $\textbf{L4b}$ multi-condition diagnosis, 0.175 \emph{versus} 0.319 (from Qwen2.5-ViT). Recall that the CLIP series were pre-trained on large-scale web data for image-text semantic matching. Hence, the CLIP features might lack fine-grained details required for discriminating dozens of fundus diseases which typically bear large inter-class similarities. Indeed we notice that for many diseases at the long tail, the LP-based classifiers built on top of the varied \ve{VE}s fail to recognize them, yielding \emph{Sensitivity} of 0 and consequently zero \emph{F1} score. As such, the performance of the \ve{VE}s is even worse than chance on \textbf{L4b}. The result suggests the limitation of pure-vision solutions for fundus image analysis.


\textbf{LLM Comparison}.
The \llm{LLM} result is shown in the \texttt{E-mode II} part of Table \ref{tab:res_overall}. The superior performance of InternVL2.5, HuatuoGPT-V and Qwen against the \ve{VE} counterpart suggests that their \llm{LLM}s  possess certain ophthalmic knowledge pertinent to fundus reading. Also notice how their performance varies over tasks, see for instance $\textbf{L4c}$ \emph{DR grading} and $\textbf{4d}$ \emph{AMD categorization}. The \llm{LLMs} perform clearly better on $\textbf{L4c}$. Our hypothesis is compared to AMD, DR-related materials are abundant online, making the \llm{LLMs} more ``familiar'' with DR. Another empirical evidence supporting this hypothesis is \llm{LLM}s' near-to-chance performance on \textbf{L2b} \emph{laterality recognition}. Such a skill is too basic to be discussed, making the related training data rare, and consequently making it a ``novel'' task for the big models. The results suggest a fundamental limitation of the current data-driven paradigm: it produces powerful models that lack basic fundus reading skills.


\textbf{MLLM Comparison}. 
The performance of the MLLMs is summarized in the last part of Table \ref{tab:res_overall}. Among the open-source models, HuatuoGPT-Vision is the best, followed by Qwen2-VL and InternVL2.5. Note that HuatuoGPT-Vision and Qwen2-VL adopt \llm{LLM} of the same structure (Qwen2-7B), yet the former's \ve{VE} (CLIP-ViT) is shown to be less effective than that of the latter (Qwen2-ViT). Such a difference shows the importance of domain-specific fine-tuning. Note that the relatively inferior performance of HuatuoGPT-Vision on localization-related tasks, see \textbf{L2a} and \textbf{L3b}. Based on our evaluation, we believe that its performance is likely to be improved when a stronger \ve{VE} is used.

In general, the MLLMs evaluated possess quite limited fundus reading skills. As shown in Table \ref{tab:corr}, the high rank correlation between MLLMs and their \llm{LLM}s clearly suggests that the former heavily rely on the latter for performing the fundus reading tasks. While both \ve{VE} and \llm{LLM} are important, the correlation analysis underscores the urgent need of developing a strong ophthalmic \llm{LLM}.

\input{tables/corr}

%% file: tables/mllm_llm_ve.tex
\begin{table}[tb!]
    \caption{\textbf{Open-source MLLMs evaluated}. Medical models are marked with *.}
    \label{tab:mllm_llm_ve}
    \renewcommand{\arraystretch}{1}
    \centerline{
        \scalebox{0.65}{ 
            \begin{tabular}{@{}l c l l @{}}
            \toprule
              \textbf{MLLM} & \textbf{HuggingFace release} &  \textbf{\ve{VE}} & \textbf{\llm{LLM}} \\ 
                \midrule
                LLaVA-v1.5-7B \cite{liu2024llava15} & 2023.10 & CLIP-ViT & Vicuna-7B \\
                *Qilin-Med-VL-Chat \cite{qilin} & 2023.12  & CLIP-ViT & Chinese-LLaMA2 \\
                LLaVA-v1.6-7B \cite{liu2024llava16} & 2024.01  & CLIP-ViT & Vicuna-7B \\ 
                *LLaVA-Med-v1.5-7B \cite{li2024llavamed}  & 2024.05 & CLIP-ViT & Mistral-7B\\
                Qwen2-VL-7B \cite{wang2024qwen2vl} & 2024.09  & Qwen2-ViT & Qwen2-7B \\
                InternVL2.5-8B \cite{internvl25} & 2024.12  & InternViT & InternLM2.5-7B \\
                *HuatuoGPT-Vision-7B \cite{chen2024huatuogpt} & 2024.06  & CLIP-ViT & Qwen2-7B \\
                Janus-Pro-7B \cite{janus} & 2025.01 & ViT-SigLIP & DeepSeek-LLM-7B \\
                Qwen2.5-VL-7B \cite{qwen2_5vl} & 2025.01  & Qwen2.5-ViT & Qwen2.5-7B \\ 
              \bottomrule
            \end{tabular}
        }
    }
\end{table}

%% file: tables/result_overall.tex
\begin{table*}[tb!]
 \caption{\textbf{Results on FunBench}. Per evaluation mode, best numbers per column are highlighted in red, while numbers lower than random guess are shown in gray. 
 }
 \label{tab:res_overall}
 \centerline{
 \renewcommand{\arraystretch}{1}
 \scalebox{0.70}{
 \begin{tabular}{@{}>{\columncolor[rgb]{1,1,1}}l lrrrrr >{\columncolor[rgb]{0.796,0.996,0.965}}r>{\columncolor[rgb]{0.796,0.996,0.965}}r>{\columncolor[rgb]{0.910,1.0,0.733}}r>{\columncolor[rgb]{0.910,1.0,0.733}}r>{\columncolor[rgb]{0.984,0.894,0.741}}r>{\columncolor[rgb]{0.984,0.894,0.741}}r>{\columncolor[rgb]{0.984,0.894,0.741}}r>{\columncolor[rgb]{0.984,0.894,0.741}}r>{\columncolor[rgb]{0.929,0.718,0.753}}r>{\columncolor[rgb]{0.929,0.718,0.753}}r>{\columncolor[rgb]{0.929,0.718,0.753}}r>{\columncolor[rgb]{0.929,0.718,0.753}}r>{\columncolor[rgb]{0.929,0.718,0.753}}r@{}}
 \toprule

\multirow{2}{*}{\cellcolor[rgb]{1,1,1}\textbf{Model}} & \multicolumn{6}{c}{\textbf{Overall performance}} & \multicolumn{2}{c}{\cellcolor[rgb]{0.796,0.996,0.965}\Gape[0pt][2pt]{\makecell{\textbf{L1:}\\ \textbf{Modality}}}} & \multicolumn{2}{c}{\cellcolor[rgb]{0.910,1.0,0.733}\Gape[0pt][2pt]{\makecell{\textbf{L2:}\\\textbf{Anatomy}}}} & \multicolumn{4}{c}{\cellcolor[rgb]{0.984,0.894,0.741}\Gape[0pt][2pt]{\makecell{\textbf{L3:}\\ \textbf{Lesion}}}} & \multicolumn{4}{c}{\cellcolor[rgb]{0.929,0.718,0.753}\Gape[0pt][2pt]{\makecell{\textbf{L4:}\\ \textbf{Disease}}}} \\
\cline{2-7}
\cellcolor[rgb]{1,1,1}& \emph{MEAN} & \emph{L1} && \emph{L2} & \emph{L3} & \emph{L4} & \emph{L1a} & \emph{L1b} & \emph{L2a} & \emph{L2b} & \emph{L3a} & \emph{L3b} & \emph{L3c} & \emph{L3d} & \emph{L4a} & \emph{L4b} & \emph{L4c} & \emph{L4d} \\

\hline

Random guess & 0.393 & 0.313 && 0.500 & 0.361 & 0.397 & 0.250 & 0.375 & 0.500 & 0.500 & 0.468 & 0.458 & 0.250 & 0.269 & 0.500 & 0.398 & 0.319 & 0.372 \\ 

\ve{E-mode I}: &&&&&&&&&&&&&&&&&& \\
CLIP-ViT & 0.578 & - && 0.820 & 0.373 & 0.541 & - & - & - & 0.820 & \color[rgb]{0.6,0.6,0.6}0.373 & - & - & - & 0.814 & \color[rgb]{0.6,0.6,0.6}0.175 & 0.556 & 0.621 \\
Qwen2-ViT & 0.597 & - && 0.892 & 0.338 & 0.560 & - & - & - & 0.892 & \color[rgb]{0.6,0.6,0.6}0.338 & - & - & - & 0.814 & \color[rgb]{0.6,0.6,0.6}0.225 & 0.506 & \color[rgb]{1,0,0}{0.694} \\
InternViT & 0.614 & - && 0.932 & 0.392 & 0.517 & - & - & - & 0.932 & \color[rgb]{0.6,0.6,0.6}0.392 & - & - & - & 0.815 & \color[rgb]{0.6,0.6,0.6}0.230 & 0.530 & 0.491 \\
ViT-SigLIP & 0.619 & - && 0.880 & 0.398 & 0.581 & - & - & - & 0.880 & \color[rgb]{0.6,0.6,0.6}0.398 & - & - & - & 0.833 & \color[rgb]{0.6,0.6,0.6}0.293 & 0.528 & 0.669 \\
Qwen2.5-ViT & 0.651 & - && 0.932 & \color[rgb]{1,0,0}{0.449} & 0.573 & - & - & - & 0.932 & \color[rgb]{1,0,0}{0.449} & - & - & - & 0.812 & \color[rgb]{0.6,0.6,0.6}0.319 & 0.514 & 0.645 \\ [2pt]
\texttt{DINOv2-large} & \color[rgb]{1,0,0}{0.655} & - && \color[rgb]{1,0,0}{0.939} & 0.408 & \color[rgb]{1,0,0}{0.616} & - & - & - & \color[rgb]{1,0,0}{0.939} & \color[rgb]{0.6,0.6,0.6}0.408 & - & - & - & \color[rgb]{1,0,0}{0.858} & \color[rgb]{1,0,0}{0.371} & \color[rgb]{1,0,0}{0.568} & 0.669\\

\llm{E-mode II}: &&&&&&&&&&&&&&&&&& \\
Janus-Pro   & 0.409 & - && \color[rgb]{1,0,0}{0.560}   & 0.213   & 0.454   & -   & -   & -   & \color[rgb]{1,0,0}{0.560} & \color[rgb]{0.6,0.6,0.6}0.213 & -   & -   & -   & -   & 0.517 & 0.446 & 0.401 \\

Qilin-Med-VL    & 0.473 & - && 0.473   & 0.519   & 0.426   & -   & -   & -   & \color[rgb]{0.6,0.6,0.6}0.473 & 0.519 & -   & -   & -   & -   & 0.552 & 0.391 & \color[rgb]{0.6,0.6,0.6}0.334 \\

LLaVA-v1.5   & 0.487 & - && 0.386   & 0.560   & 0.515   & -   & -   & -   & \color[rgb]{0.6,0.6,0.6}0.386 & 0.560 & -   & -   & -   & -   & 0.521 & 0.557 & 0.467 \\

LLaVA-Med-v1.5   & 0.529 & - && 0.483   & 0.507   & 0.597   & -   & -   & -   & \color[rgb]{0.6,0.6,0.6}0.483 & 0.507 & -   & -   & -   & -   & 0.638 & 0.640 & 0.513 \\

LLaVA-v1.6  & 0.604 & - && 0.502   & 0.717   & 0.594   & -   & -   & -   & 0.502 & 0.717 & -   & -   & -   & -   & 0.606 & 0.647 & 0.529 \\

Qwen2.5-VL    & 0.654 & - && 0.497   & 0.735   & 0.729   & -   & -   & -   & \color[rgb]{0.6,0.6,0.6}0.497 & 0.735 & -   & -   & -   & -   & 0.757 & 0.726 & 0.704 \\

Qwen2-VL        & 0.673 & - && 0.488   & 0.747   & 0.783   & -   & -   & -   & \color[rgb]{0.6,0.6,0.6}0.488 & 0.747 & -   & -   & -   & -   & 0.750 & 0.855 & 0.743 \\

HuatuoGPT-V & 0.692 & - && 0.541   & 0.706   & \color[rgb]{1,0,0}{0.829}   & -   & -   & -   & 0.541 & 0.706 & -   & -   & -   & -   & \color[rgb]{1,0,0}{0.793} & 0.927 & \color[rgb]{1,0,0}{0.768} \\

InternVL2.5  & \color[rgb]{1,0,0}{0.703} & - && 0.488   & \color[rgb]{1,0,0}{0.831}   & 0.789   & -   & -   & -   & \color[rgb]{0.6,0.6,0.6}0.488 & \color[rgb]{1,0,0}{0.831} & -   & -   & -   & -   & 0.792 & \color[rgb]{1,0,0}{0.929} & 0.647 \\

\texttt{E-mode III}: &&&&&&&&&&&&&&&&&& \\
Janus-Pro  & \color[rgb]{0.6,0.6,0.6}0.355 & 0.777 && \color[rgb]{0.6,0.6,0.6}0.167 & \color[rgb]{0.6,0.6,0.6}0.108 & \color[rgb]{0.6,0.6,0.6}0.369 & 0.943 & 0.611 & \color[rgb]{0.6,0.6,0.6}0.189 & \color[rgb]{0.6,0.6,0.6}0.145 & \color[rgb]{0.6,0.6,0.6}0.115 & \color[rgb]{0.6,0.6,0.6}0.203 & \color[rgb]{0.6,0.6,0.6}0.060 & \color[rgb]{0.6,0.6,0.6}0.055 & 0.523 & \color[rgb]{0.6,0.6,0.6}0.319 & \color[rgb]{0.6,0.6,0.6}0.278 & \color[rgb]{0.6,0.6,0.6}0.358 \\

LLaVA-v1.5 & 0.418 & 0.722 && \color[rgb]{0.6,0.6,0.6}0.489 & \color[rgb]{0.6,0.6,0.6}0.165 & \color[rgb]{0.6,0.6,0.6}0.296 & 0.974 & 0.470 & \color[rgb]{0.6,0.6,0.6}0.492 & \color[rgb]{0.6,0.6,0.6}0.485 & \color[rgb]{0.6,0.6,0.6}0.276 & \color[rgb]{0.6,0.6,0.6}0.266 & \color[rgb]{0.6,0.6,0.6}0.013 & \color[rgb]{0.6,0.6,0.6}0.104 & \color[rgb]{0.6,0.6,0.6}0.364 & \color[rgb]{0.6,0.6,0.6}0.292 & \color[rgb]{0.6,0.6,0.6}0.187 & \color[rgb]{0.6,0.6,0.6}0.341 \\

Qilin-Med-VL & 0.426 & 0.651 && \color[rgb]{0.6,0.6,0.6}0.496 & \color[rgb]{0.6,0.6,0.6}0.211 & \color[rgb]{0.6,0.6,0.6}0.347 & 0.910 & 0.392 & 0.504 & \color[rgb]{0.6,0.6,0.6}0.489 & \color[rgb]{0.6,0.6,0.6}0.463 & \color[rgb]{0.6,0.6,0.6}0.361 & \color[rgb]{0.6,0.6,0.6}0.000 & \color[rgb]{0.6,0.6,0.6}0.019 & \color[rgb]{0.6,0.6,0.6}0.491 & \color[rgb]{0.6,0.6,0.6}0.354 & \color[rgb]{0.6,0.6,0.6}0.214 & \color[rgb]{0.6,0.6,0.6}0.328 \\

LLaVA-v1.6 & 0.435 & 0.777 && \color[rgb]{0.6,0.6,0.6}0.449 & \color[rgb]{0.6,0.6,0.6}0.194 & \color[rgb]{0.6,0.6,0.6}0.319 & 0.970 & 0.585 & \color[rgb]{0.6,0.6,0.6}0.400 & \color[rgb]{0.6,0.6,0.6}0.497 & \color[rgb]{0.6,0.6,0.6}0.461 & \color[rgb]{0.6,0.6,0.6}0.243 & \color[rgb]{0.6,0.6,0.6}0.058 & \color[rgb]{0.6,0.6,0.6}0.013 & \color[rgb]{0.6,0.6,0.6}0.357 & \color[rgb]{0.6,0.6,0.6}0.314 & \color[rgb]{0.6,0.6,0.6}0.289 & \color[rgb]{0.6,0.6,0.6}0.315 \\

LLaVA-Med-v1.5 & 0.440 & 0.813 && 0.507 & \color[rgb]{0.6,0.6,0.6}0.215 & \color[rgb]{0.6,0.6,0.6}0.223 & 0.961 & 0.665 & 0.502 & 0.512 & \color[rgb]{0.6,0.6,0.6}0.166 & \color[rgb]{0.6,0.6,0.6}0.349 & 0.261 & \color[rgb]{0.6,0.6,0.6}0.086 & \color[rgb]{0.6,0.6,0.6}0.000 & \color[rgb]{0.6,0.6,0.6}0.255 & \color[rgb]{0.6,0.6,0.6}0.290 & \color[rgb]{0.6,0.6,0.6}0.345 \\

Qwen2.5-VL  & 0.480 & 0.942 && \color[rgb]{0.6,0.6,0.6}0.364 & \color[rgb]{0.6,0.6,0.6}0.231 & \color[rgb]{0.6,0.6,0.6}0.384 & 0.978 & 0.906 & \color[rgb]{0.6,0.6,0.6}0.310 & \color[rgb]{0.6,0.6,0.6}0.417 & \color[rgb]{0.6,0.6,0.6}0.386 & \color[rgb]{0.6,0.6,0.6}0.242 & \color[rgb]{0.6,0.6,0.6}0.218 & \color[rgb]{0.6,0.6,0.6}0.077 & 0.532 & \color[rgb]{0.6,0.6,0.6}0.394 & \color[rgb]{0.6,0.6,0.6}0.303 & \color[rgb]{0.6,0.6,0.6}0.308 \\

InternVL2.5 & 0.506 & 0.946 && \color[rgb]{1,0,0}{0.564} & \color[rgb]{0.6,0.6,0.6}0.244 & \color[rgb]{0.6,0.6,0.6}0.269 & \color[rgb]{1,0,0}{0.997} & 0.894 & \color[rgb]{1,0,0}{0.640} & \color[rgb]{0.6,0.6,0.6}0.488 & \color[rgb]{0.6,0.6,0.6}0.391 & \color[rgb]{0.6,0.6,0.6}0.144 & \color[rgb]{0.6,0.6,0.6}0.227 & \color[rgb]{0.6,0.6,0.6}0.213 & \color[rgb]{0.6,0.6,0.6}0.031 & 0.400 & \color[rgb]{0.6,0.6,0.6}0.294 & \color[rgb]{0.6,0.6,0.6}0.351 \\

Qwen2-VL & 0.520 & 0.925 && \color[rgb]{0.6,0.6,0.6}0.497 & \color[rgb]{0.6,0.6,0.6}0.316 & \color[rgb]{0.6,0.6,0.6}0.343 & 0.995 & 0.855 & 0.506 & \color[rgb]{0.6,0.6,0.6}0.488 & 0.503 & \color[rgb]{1,0,0}{0.432} & \color[rgb]{0.6,0.6,0.6}0.073 & \color[rgb]{1,0,0}{0.255} & 0.508 & \color[rgb]{0.6,0.6,0.6}0.352 & \color[rgb]{0.6,0.6,0.6}0.202 & \color[rgb]{0.6,0.6,0.6}0.311 \\

HuatuoGPT-V & 0.523 & 0.933 && \color[rgb]{0.6,0.6,0.6}0.309 & \color[rgb]{1,0,0}{0.374} & \color[rgb]{1,0,0}{0.477} & 0.961 & 0.905 & \color[rgb]{0.6,0.6,0.6}0.140 & \color[rgb]{0.6,0.6,0.6}0.477 & \color[rgb]{1,0,0}{0.548} & \color[rgb]{0.6,0.6,0.6}0.379 & 0.338 & \color[rgb]{0.6,0.6,0.6}0.231 & \color[rgb]{1,0,0}{0.622} & \color[rgb]{1,0,0}{0.504} & 0.414 & \color[rgb]{0.6,0.6,0.6}0.369 \\

\texttt{GPT-4o} & \color[rgb]{1,0,0}{0.542} & \color[rgb]{1,0,0}{0.961} && 0.535 & \color[rgb]{0.6,0.6,0.6}0.341 & \color[rgb]{0.6,0.6,0.6}0.331 & 0.965 & \color[rgb]{1,0,0}{0.957} & 0.557 & \color[rgb]{1,0,0}{0.514} & \color[rgb]{0.6,0.6,0.6}0.362 & \color[rgb]{0.6,0.6,0.6}0.412 & \color[rgb]{1,0,0}{0.361} & \color[rgb]{0.6,0.6,0.6}0.228 & \color[rgb]{0.6,0.6,0.6}0.018 & 0.452 & \color[rgb]{1,0,0}{0.476} & \color[rgb]{1,0,0}{0.378} \\




\bottomrule
\end{tabular}}}
\end{table*}

%% file: tables/corr.tex
\begin{table}[tb!]
    \caption{\textbf{Correlation analysis in terms of MEAN-performance ranks}.}
    \label{tab:corr}
    \renewcommand{\arraystretch}{1}
    \centerline{
        \scalebox{0.8}{ 
            \begin{tabular}{@{}ll r @{}}
            \toprule
              \textbf{Module} && \textbf{Spearman-correlation to MLLM}  \\ 
 \midrule
                LLM && 0.917\\
                VE && 0.055\\

              \bottomrule
            \end{tabular}
        }
    }
\end{table}